\documentclass[runningheads]{llncs}
\usepackage{graphicx}
\usepackage{amsmath}
\usepackage{listings}
\usepackage{rotating}
\usepackage{booktabs}
\usepackage{makecell}
\usepackage{xcolor}
\usepackage{soul}
\usepackage{todonotes}
\usepackage{comment}

\definecolor{codegreen}{rgb}{0,0.6,0}
\definecolor{codegray}{rgb}{0.5,0.5,0.5}
\definecolor{codepurple}{rgb}{0.58,0,0.82}
\definecolor{backcolour}{rgb}{0.95,0.95,0.92}
\lstdefinestyle{mystyle}{
    backgroundcolor=\color{backcolour},
    belowcaptionskip=1\baselineskip,
    showstringspaces=false,
    basicstyle=\footnotesize\ttfamily,
    commentstyle=\itshape\color{green!40!black},
    identifierstyle=\color{blue},
    stringstyle=\color{orange},
}
\begin{document}
\title{The Emotion Frame Ontology} 
%
%
\author{Stefano De Giorgis\inst{1}\orcidID{0000-0003-4133-3445} \and
Aldo Gangemi\inst{2,3}\orcidID{0000-0001-5568-2684}}
\authorrunning{De Giorgis et al.}
%
%
\maketitle              
\begin{abstract}
Emotions are a subject of intense debate in various disciplines. Despite the proliferation of theories and definitions, there is still no consensus on what emotions are, and how to model the different concepts involved when we talk about –– or categorize –– them.
In this paper, we propose an OWL frame-based ontology of emotions: the Emotion  Frames Ontology (EFO).  
EFO treats emotions as semantic frames, with a set of semantic roles that capture the different aspects of emotional experience. 
EFO follows pattern-based ontology design, and is aligned to the DOLCE foundational ontology. 
EFO is used to model multiple emotion theories, which can be cross-linked as modules in an Emotion Ontology Network. In this paper, we exemplify it by modeling Ekman's Basic Emotions (BE) Theory as an EFO-BE module, and demonstrate how to perform automated inferences on the representation of emotion situations.
EFO-BE has been evaluated by lexicalizing 
the BE emotion frames from within the Framester knowledge graph, and implementing a graph-based emotion detector from text. In addition, an EFO integration of multimodal datasets, including emotional speech and emotional face expressions, has been performed to enable further inquiry into cross-modal emotion semantics. 

\end{abstract}
\section{Introduction}

Emotions are subject of intense debate in various disciplines from philosophy \cite{deonna2012emotions}, neuroscience \cite{phillips2003neurobiology}, economics \cite{loewenstein2000emotions}, biology \cite{dalgleish2004emotional}, developmental psychology \cite{plutchik2003emotions}, sociology \cite{turner2005sociology}, and cultural anthropology \cite{lutz1986anthropology}. Despite the proliferation of theories and definitions, there is still no consensus on what emotions are, and how to model the different concepts involved when we talk about –– or categorize –– them. Since different theories focus on different aspects of emotion situations, we have designed an OWL ontology network to jointly represent those aspects, in order to integrate heterogeneous theories or emotion datasets.

In addition, multimodal data applied to emotion research may address heterogeneous entities. For example, electrodermal activity (EDA) biofeedback data (addressing skin conductance, heat, biochemical markers such as sweating, etc.) assume emotions as physiological states, but those states can be associated with internal emotions, with an appraisal of others' emotions, or even with an attempt to raise emotions in other people.

Furthermore, beyond emotions as physiological states, we can categorise them as social concepts, as behaviors, as attitudes towards what happens within one's mind, or outside of it, etc.

Finally, emotions are often overlapping or mixing with feelings, moods, or other affective situations.

We have designed the Emotion Frame Ontology (EFO) in order to integrate the models and the data about those different aspects of emotions, and to perform automated reasoning and semantically-informed machine learning on them.
EFO represents the concept of ``emotion'' as a semantic frame, intended as in Fillmore's frame semantics \cite{fillmore1982framsemantics}.
Frames are schematizations of recurrent situations, formalising knowledge in terms of semantic roles (frame elements) participating to a certain world situation  (frame occurrence).
EFO is modeled following the eXtreme Design methodology \cite{blomqvist2010experimenting}, it reuses ontology design patterns (ODPs) \cite{presutti2008content} and it is aligned to the DOLCE \cite{gangemi2002sweetening}, \cite{borgo2022dolce} (Descriptive Ontology for Linguistic and Cognitive Engineering) foundational ontology.
The OWL version of DOLCE (DOLCE-Zero\footnote{\url{http://www.ontologydesignpatterns.org/ont/d0.owl}}) is used because it fits the cognitive aspects considered in this work, and includes the Description\&Situation \cite{gangemi2003understanding} ODP, 
which enables a reified relational modelling of an emotion as both a \texttt{:ConceptualFrame}, and its occurrences as emotion situations (as an \texttt{:EmotionSituation}).

Emotion situations on their turn may include physiological manifestations (e.g. ``blushing'', ``accelerated heart beat'', ``becoming pale'', etc.) as a \newline \texttt{:PhysicalManifestation}. Emotion situations can be linked to multiple agents that produce, recognise, evaluate, or attempt at activating physiological manifestations of emotions. This broad conceptualisation of emotions is ideal to model them as a hub of different types of knowledge (factual, opinionated, perspectival, etc.), and across multimodal manifestations (textual, visual, auditory, olfactory).

As a consequence, EFO can model multiple emotion theories at the same time, treating the theoretical concept of ``emotion'', as intended in each theory, as a projection of the general \texttt{efo:Emotion} frame. 
For example, EFO integrates cognitive-centered theories focused on the nature of emotions (e.g., Plutchik's Emotion Wheel \cite{plutchik1980emotion,plutchik1982psychoevolutionary}), with socio-behavioral-centered models (e.g., the appraisal models inspired by Ortony, Clore and Collins    \cite{ortony1988emotion,ortony2022cognitive}). This projection-based integration operates at the intensional level of frames (only some roles of the frames are activated when using one theory or another), and is reflected in cognitive vs. socio-behavioural situations resulting as parts of a same emotion situation.


\section{Related Work}
\label{sec:related_work}


\subsection{Emotion Ontologies}
\label{subsec:emotions_formalization}

Although not in a large number, some ontologies have attempted to formalize the domain of emotions; they are briefly presented here.

\paragraph{\textbf{EmOCA}}
Emotion Ontology for Context Awareness \cite{berthelon2013emotion} is an ontology developed to improve emotion detection from physiological manifestation, considering contextual knowledge such as ``X has a certain attitude towards Y''. The main aim is to infer \textit{phobia/philia} attitudes. It uses RDF and RDFS (but not OWL), its main classes are \texttt{:Stimulus} and \texttt{:PersonalityTrait}. Finally, it adopts Ekman's Basic Emotions theory\footnote{The EmOCA ontology is available at \url{http://ns.inria.fr/emoca/}}.

\paragraph{\textbf{EmotionsOnto}}
EmotionsOnto \cite{lopez2014emotionsonto} is a general emotion ontology importing DOLCE and reusing the Description\&Situation (D\&S) ODP. It is written in OWL and it includes the possibility to introduce multimodal data from expression and sensory-based recognition systems. Among the emotion ontologies here presented this is the closest to our work, since it also reuses the notion of FrameNet frames.
The main improvements of EFO over this work are (i) lexicalization of emotion frames, as described in Sect. \ref{sec:bet}, (ii) the reuse of semantic web resources, (iii) the ability to model multiple theories simultaneously, and (iv) the reuse of existing multimodal datasets transposed to RDF format.

\paragraph{\textbf{SOCAM Affective Extension}}
Ontology-based Affective Context Representation \cite{benta2007ontology}, based on the SOCAM framework \cite{wang2004ontology} represents affective states for context aware applications. It is written in OWL, and introduces fuzzy logics to reason over more complex affective states.

\paragraph{\textbf{MFOEM}} 
Developed by Hastings et al. \cite{hastings2011emotion}, EMO or MFOEM is the ontological module dedicated to emotions in OBO Foundry \cite{smith2007obo}, based on the BFO foundational ontology \cite{arp2015building}. It is developed based on the emotional theory described by Sanders and Scherer \cite{sander2009oxford}.
In the MFOEM ontology,  
the concept of ``emotion'' is represented as an \texttt{mfoem:emotion process}\footnote{Since MFOEM uses codes as class names, 
we use here an 'anomalous' notation, providing the 'mfoem:' prefix followed by the class label.},
declared as a subclass of an \texttt{mfoem:affective process}.
All emotions are listed as subclasses of \texttt{mfoem:emotion process}.
MFOEM includes relevant classes such as \texttt{mfoem:physiological process involved in an} \texttt{emotion}, described as ``A bodily process that encompasses all the neurophysiological changes accompanying an emotion, which take place in the central nervous system (CNS), neuro-endocrine system (NES) and autonomous nervous system (ANS).''. 
The major differences between EFO and MFOEM include: (i) EFO represents multiple emotion theories at the same time (e.g. Ekman, Plutchik, OCC, etc.); (ii) emotions are represented as semantic frames; and (iii) EFO commits to DOLCE instead of BFO. Since DOLCE is not as committed to philosophical struggles about ``objective reality'' as BFO does, it is more appropriate for data that contain socially-constructed entities such as frame, roles, concepts, etc., which are widespread in cognitive and social sciences
\footnote{For those interested in the longstanding debate between DOLCE and BFO foundations, cf. the Joint Ontology Workshop 2022 (JOWO 2022) special FOUST (Foundational Ontologies Workshop) event ``The Foustian struggle'' at 
\url{https://www.iaoa.org/jowo/2022/foustian\_struggle/index.html}}.


\paragraph{\textbf{framECO}} 

framECO, developed by Coppini et al. \cite{coppini2022frameco}, adopts a transversal approach to represent emotion content in literary works. It is frame-based, and introduces specific classes to represent non-trivial emotion situations, taking inspiration in particular from the Dictionary of Obscure Sorrows \cite{koenig2021dictionary}, a repository of ``unfamiliar emotions'' such as ``\textit{Onism}: n. the frustration of being stuck in just one body, that inhabits only one place at a time [...].'' It consists of a new, integrated emotion theory 
formalized in OWL \footnote{\url{https://saroppini.github.io/framECO/}}. framECO is also based on experimental cognitive results about the difficulty of categorising non-trivial emotion situations with current emotion theories such as those presented in Sect. \ref{sec:emotion_theories}.

\paragraph{\textbf{WordNet Affect}}
WordNet-Affect \cite{strapparava2004wordnet} is not a proper ontology, but a WordNet extension including emotion or valence annotations for WordNet synsets. Synsets are labeled according to four labels: positive, negative, ambiguous, and neutral. The semantic schema of WordNet-Affect includes the stative/causative dimension: an adjective is ``causative'' if it can be used to evoke an emotion \texttt{:causedBy} the
entity whose word is modified by the adjective, as in `amazing roadtrip'. An adjective is ``stative'' if it refers to an emotion owned or felt by the agent whose word is modified the adjective, as in `caring/joyful person'. WordNet-Affect has been the first of several lexical resources (e.g., SenticNet, DepecheMood, etc.) that categorize words or word senses according to the emotions that can be evoked by them: they are not covered in this quick state of the art because of their informality, and because they use simple data models that do not make particular assumptions about emotion evocation.

\subsection{Emotions theories}
\label{sec:emotion_theories}

Emotion theories are generally distinguished in two main areas: (i) Categorical models and (ii) Dimensional models.
Dimensional models usually represent emotions in a continuous space over two dimensions: arousal and polarity. In this work we focus on categorical models, planning to include dimensional ones as future work.
Categorical models (as well as appraisal theories) organise emotions as discrete categories, possibly introducing some form of taxonomy, but mainly focusing on a small group of concepts, often labeled ``basic emotions'' or ``primary emotions''.
Depending on the categorical model adopted, basic emotions can vary in number: 5 in Ekman's theory \cite{ekman1999basic,ekman2018atlas}, 6 in Mehrabian \cite{mehrabian1997comparison} and 7 in Izard \cite{izard1971face}. In this work we focus on Ekman's Basic Emotions theory. 

Ekman's theory proposes 6 ``Basic Emotions'': Enjoyment, Curiosity, Fear, Sadness, Anger, and Disgust, as reported in the 2021 version \cite{ekman2018atlas}, available at \url{atlasofemotions.org}. 
Since the Ekman's EFO module is obtained by reverse engineering the Atlas of Emotions website, details of the BE theory are given in Sect. \ref{sec:be}, describing the ontology.
The BE theory is the first theory introduced in EFO due to its pervasiveness, both from a theoretical perspective, and from an operational one, thanks to its reuse in tasks of emotion detection from text and emotion recognition from facial expression, using the Ekman's Facial Action Coding System (FACS) \cite{ekman1978facial}.
For a detailed survey on emotion recognition techniques and datasets, refer to Canal et al. \cite{canal2022survey}.

\section{Emotion Frame Ontology}
\label{sec:efo}

The Emotion Frame Ontology (EFO) is an ontology network for representing emotions as frames, according to different emotion theories. Adopting a formal frame structure, we tackle the problem of ``what is an emotion'': there is some cognitive entity resulting from reifying a relation including physiological states that are experienced, recognised, appraised, or provoked by agents in their interaction as an internal cognitive state, or as a environmental or social situation, in some spatio-temporal context. Such entities are expressed by information objects (texts, pictures, sounds, odours, etc.)

We are therefore talking about emotions as socio-cognitive situations existing at some spatio-temporal point, and dependent on internal states of agents, their dynamic interaction in complex environments, as well as on the information used to conceptualize those states, or the overall situation.

Since emotion situations are observed as complex relations, we need to reify them as individual situations. And since different theories see different parts of those relations, we also need to talk about what parts of those relations are used in what theory. 
This requirement calls for a mixed intensional-extensional representation domain, which in OWL can be formalised by using punning. We adopt the frame pattern as formalised in Framester \cite{gangemi2016framester} (modeled on its turn on the Description\&Situation ODP), which allows to jointly talk about the roles and types of an emotion frame, and about the classes of entities that play those roles (physiological states, agents, space, time, social scenarios, information objects, etc.).

This notion of emotion contextual to a certain theory is modeled as a frame \texttt{rdfs:subClassOf} the most general \texttt{efo:Emotion} conceptual frame, and it takes as roles the specific dimensions and aspects covered in that theory.

The EFO ontology network is constituted by: (i) the Emotion Core (EmoCore) module, (ii) the module formalising Ekman's Basic Emotion (BE) theory from the Atlas of Emotions\footnote{ \url{https://atlasofemotions.org/}}, and (iii) a module of Basic Emotions triggers (BET), containing lexical and factual triggers extracted from semantic web resources, in order to operationalise EFO as a fully explainable graph-based emotion detector. 
The entities declared as emotion triggers are extracted by querying the Framester ontological hub \cite{gangemi2016framester,gangemi2020closing}, described in the next paragraph.

\paragraph{\textbf{Framester Ontology}}
Framester provides a formal semantics for semantic frames in a curated linked data version including multiple linguistic and factual data resources. It is based on the formal representation of frames from FrameNet \cite{nuzzolese2011gathering}. Several resources are integrated in the hub besides FrameNet: WordNet \cite{miller1998wordnet}, VerbNet \cite{schuler2005verbnet}, a cognitive layer including MetaNet \cite{gangemi2018amnestic} and ImageSchemaNet \cite{de2022imageschemanet}, multilingual resources such as BabelNet \cite{navigli2010babelnet}, factual knowledge bases (e.g. DBpedia \cite{auer2007dbpedia}, YAGO \cite{suchanek2007yago}, etc.), and ontology schemas (e.g. DOLCE in its versions \cite{borgo2022dolce,gangemi2003sweetening}), with formal links between them, resulting in a strongly connected RDF/OWL knowledge graph.
Framester can be used to jointly query the above mentioned resources via its SPARQL endpoint\footnote{Framester endpoint is available here: \url{http://etna.istc.cnr.it/framester2/sparql}}.

EFO modules are available on the EFO GitHub\footnote{https://anonymous.4open.science/r/EFO-C124/EmoCore\_iswc.ttl}. A complete version will be made available and queryable from a stable SPARQL endpoint at camera-ready time.

\subsection{EmoCore Module}
\label{sec:emocore}

The EmoCore module contains the minimal vocabulary to talk about emotion relations. It models the notion of ``emotion'' as a frame. It is aligned to the DOLCE foundational ontology, and reuses the Description\&Situation ontology design pattern \cite{gangemi2003understanding,gangemi2008norms}, to express emotion both as classes of situation and as conceptual individuals (in OWL2 syntax).
Each emotion theory (already included, or foreseen in future developments) is modeled in a separate module importing EmoCore. The class \texttt{efo:Emotion} is a subclass of \texttt{fs:ConceptualFrame}, reused from Framester, in turn subclass of \texttt{dul:Description}, from the DOLCE-Zero foundational ontology. 
Each emotion description can be satisfied by some \texttt{efo:EmotionSituation}, namely, the realization/occurrence of an event or state involving some emotion. Note that an emotion situation may involve entities of a different type: a mental state, the expression of some appraisal consequence of some emotion state, the triggering moment of some emotion, etc. 
Being a core module, it generalises specific notions of emotions, in order to cover every possible emotion situation. The aim of such a module is to have a broad \texttt{efo:Emotion} class  as superclass for encompassing all the possible emotion concepts, as assumed by all theories.

\paragraph{\textbf{EmoCore Classes}} 
The main class of the EmoCore module is the \texttt{efo:Emotion} class: this class is the broadest possible notion of ``emotion''. It is meant to encompass all theory-specific definitions. It also encompasses lexical definitions (formalised as classes in Framester), as with 
the WordNet synsets: \newline \texttt{wn:synset-emotion-noun-1} and \texttt{wn:synset-emotional\_state-noun-1}, which contribute the many lexical concepts/frames evoking emotion situations.
It is subclass of \texttt{fs:ConceptualFrame}, which in turn is subClassOf \texttt{dul:Description}. Under the \texttt{dul:Situation} class we put the realisation of any emotion situation, namely an emotion \texttt{fs:FrameOccurrence}. 
Its subclasses also include FrameNet frames that cover emotional phenomenon from different perspectives:
\begin{itemize}
    \item \texttt{fs:EmotionActive} focuses on the positive or negative push that an Undergoer feels from the emotion: the focus is on the ``activity'' that an emotion exerts on a subject;
    \item \texttt{fs:EmotionDirected} focuses on describing an Experiencer, who is feeling or experiencing a particular emotional response to a Stimulus, or about a Topic
    \item \texttt{fs:Feeling} focuses on representing an emotional state, which may involve an appraisal of the emotional state;
    \item \texttt{fs:MentalProperty} is a general frame to refer to any possible mental state, being it known or inferred from a behavioral manifestation 
\end{itemize}

\paragraph{\textbf{EmoCore Properties}} 
A `triggering' binary relation seems shared across all different emotion theories, and is represented in EFO as a \texttt{efo:triggers} object property, used to declare some entity as trigger of an emotion.

\subsection{BE Module}
\label{sec:be}

The Basic Emotions module is the formalisation of Basic Emotions theory \cite{ekman1999basic} 
\footnote{The current Ekman theory is available at: \url{https://atlasofemotions.org}}, with the addition of the ``curiosity'' emotion, from its original version. 
Besides the six basic emotions, the current version of Ekman’s theory includes other aspects, such as ``Mood'' or ``Pre-condition'', namely those internal or external states that influence the emergence of a certain emotion as a consequence of some stimulus. The main competency questions that can be answered in this module are:

\begin{enumerate}
    \item \textbf{CQ1}: what and how many emotions are in Ekman’s theory?
    \item \textbf{CQ2}: what is the polarity of each Basic Emotion?
    \item \textbf{CQ3}: what psychopathology makes a subject tend to what emotion?
    \item \textbf{CQ4}: what is an ``emotion counter'' or ``emotion antidote'' and what are the emotion counters/antidotes for some specific emotion?
    \item \textbf{CQ5}: which emotion is more intense than another?
\end{enumerate}

\paragraph{\textbf{BE Classes}}

Here we list some of the main classes in the BE module:
\begin{itemize}
    \item \texttt{be:PreCondition}: the context or situation that may influence the way a subject enters the emotion.
    \item \texttt{be:BE\_Emotion}: The class \texttt{be:BE\_Emotion} is the class to represent those entities that are said to be ``emotions'' in the Basic Emotions theory. Being primitive they are not provided a definition. Each emotion takes as subclasses more specific states organized on an increasing state of intensity.
    \item \texttt{be:EmotionCounter}: some counterforce to some emotional state. If the emotion is positive the counter is a \texttt{be:EmotionImpediment}, namely some other emotional state conflicting with the positive one. If the emotion is negative, the counter is an \texttt{be:EmotionAntidote}: usually an action requiring some intentional commitment (e.g. the \texttt{be:AnxietyAntidote}) is defined as ``Making a special effort of letting go of ruminations about the past and anticipations of the future''.
    \item \texttt{be:Mood}: longer lasting emotional states, they foster the repetition of the same emotional state even without explicit trigger.
    \item \texttt{be:PerceptionDatabase}: the set of universal/hardwired responses and individually acquired emotional memories. This set has an influence on the appraisal process performed on the trigger, and therefore can influence the deriving emotional state.
    \item \texttt{be:Trigger}: the interaction between the appraisal and some hard-wired/acquired script in the Perception Database.
    \texttt{be:PersonalityTrait}: a tendency that make a person lean more often towards certain emotional states.
    \item \texttt{be:PhysicalChange}: the change in our body when some emotion arises.
    \item \texttt{be:PhysiologicalChange}: the qualitative experience of the emotion, namely the manifestation that some emotion determines.
    \item \texttt{be:Psychopathology}: pathologies that can be traced back to some emotion, each pathology has a prototypical \texttt{be:emotionalTendencyTowards} some emotional state.
    \item \texttt{be:SelectiveFilterPeriod}: the state that, given a certain initial trigger of e.g. fear, determines a narrowed and distorted perception, filtering and interpreting information consistent to the prevailing emotion.
    \item \texttt{be:Signal}: external universal prototypical manifestations of some emotion displayed via facial expressions or voice tone.
    \item \texttt{be:PostCondition}: the result of an emotional action. It can be both external or internal. The post condition can drive to a following emotional state.
\end{itemize}

\paragraph{\textbf{BE Properties}}
To represent possible semantic relations among classes participating in an emotion situation, according to Basic Emotions theory, we list here the main object properties in the BE module:

\begin{itemize}
    \item \texttt{be:emotionalTendencyTowards}: some \texttt{be:Psychopathology} make a subject tend to some \texttt{be:BE\_Emotion};
    \item \texttt{be:hasAntidote}: some negative emotion has as emotion counter some \newline \texttt{be:EmotionAntidote};
    \item \texttt{be:hasImpediment}: some positive emotion has as emotion counter some \texttt{be:EmotionImpediment};
    \item \texttt{be:hasPreCondition}: some emotional state has pre-condition some \newline \texttt{be:PreCondition}, to be taken in consideration in the final \texttt{be:PostCondition} state;
    \item \texttt{be:moreIntenseThan} and \texttt{be:lessIntenseThan}: some subclass of some \newline  \texttt{be:BE\_Emotion} is more or less intense of some other subclass.
\end{itemize}

The Competency Questions presented in this section can be answered by querying the BE ontological module. The following query answer to the BE competency questions, reusing classes and properties described in the paragraphs above, in order to investigate the ontological nature of emotions as modeled in Ekman’s Basic Emotions theory.

\begin{lstlisting}[basicstyle=\fontsize{7}{8}\selectfont\ttfamily,aboveskip=0pt,belowskip=0pt, belowcaptionskip=0pt, captionpos=b, label=lst:sparql_query, caption= {SPARQL Query to explore the Basic Emotions ontological module.}, label=lst:datarequest,frame=single,linewidth=1\columnwidth,breaklines=true,language=SPARQL, float=ht]

SELECT DISTINCT ?emotion ?polarity ?psychopathology ?subEmotion ?antidote
?action
WHERE {
?emotion rdfs:subClassOf be:BE_Emotion .
?emotion be:hasPolarity ?polarity .
?psychopathology be:emotionalTendencyTowards ?emotion . 
?emotion be:hasPersonalityTrait ?personalityTrait . 
?subEmotion rfds:subClassOf ?emotion ;
be:hasAntidote|be:hasImpediment ?antidote ;
be:moreIntenseThan ?siblingEmotion .
?emotion be:hasAction ?action. 

FILTER(regex(str(?emotion), 'Fear')) 
}

\end{lstlisting}

The query, explained in natural language investigate BE graph to retrieve all emotions, filtered by containing in their string ``fear'', therefore the example that we provide
here includes knowledge extracted for the \texttt{be:Fear} emotion. The first triple requested is for the entity ?emotion to be subclass of the \texttt{be:BE\_Emotion} class. The only entity, subclass of the general emotion class in BE module, containing in its string ``fear'' is \texttt{be:Fear}. Therefore, it asks for the \texttt{be:Fear} polarity, which is \texttt{be:NegativePolarity},
and all the psychopathologies whose patients tend towards {be:Fear}, which in BE are: \texttt{be:AvoidantPersonalityDisorder}, \texttt{be:GeneralizedAnxietyDisorder}, \texttt{be:ObsessiveCompulsiveDisorder}, \newline \texttt{be:PostTraumaticStressDisorder}, and \texttt{be:SocialAnxietyDisorder}. \newline
The \texttt{be:Fear} personality trait is \texttt{be:FearPersonality}, described as ``A shy or timid person. This personality type is likely to avoid risks and uncomfortable situations. Timid people may perceive the world as full of difficult situations''. 
Its subclasses are \texttt{be:Anxiety}, \texttt{be:Desperation}, \texttt{be:Dread}, \texttt{be:Horror}, \texttt{be:Nervousness}, \texttt{be:Panic}, \texttt{be:Terror}, and \texttt{be:Trepidation}; but not all of them are declared as having some ?antidote. Respectively, those retrieved by this query are \texttt{be:AnxietyAntidote}, \texttt{be:DreadAntidote}, \texttt{be:HorrorAntidote}, \texttt{be:NervousnessAntidote}, \texttt{be:PanicAntidote}, and \texttt{be:TrepidationAntidote}. Finally, the list of sub-emotions for \texttt{be:Fear}, ordered for increasing intensity are: Trepidation $<$ Nervousness $<$ Anxiety $<$ Dread $<$ Desperation
$<$ Panic $<$ Horror $<$ Terror; where the ``$<$'' symbol here stands for the \texttt{be:moreIntenseThan} object property.

The ontology has proven consistent via testing it with Hermit $1.4.3.456$ reasoner in Protégé, version $5.5.0$.

\section{Operationalizing EFO}
\label{sec:operationalizing_efo}

To operationalize the EFO ontology, we firstly populated each Ekman's Basic Emotion knowledge graph with triggers from well known semantic web resources, then we developed a graph based emotion detector from natural language.

\subsection{BE Triggers Module Use Case: Disgust}
\label{sec:bet}

\begin{sidewaysfigure}
  \centering
  \includegraphics[scale=0.25]{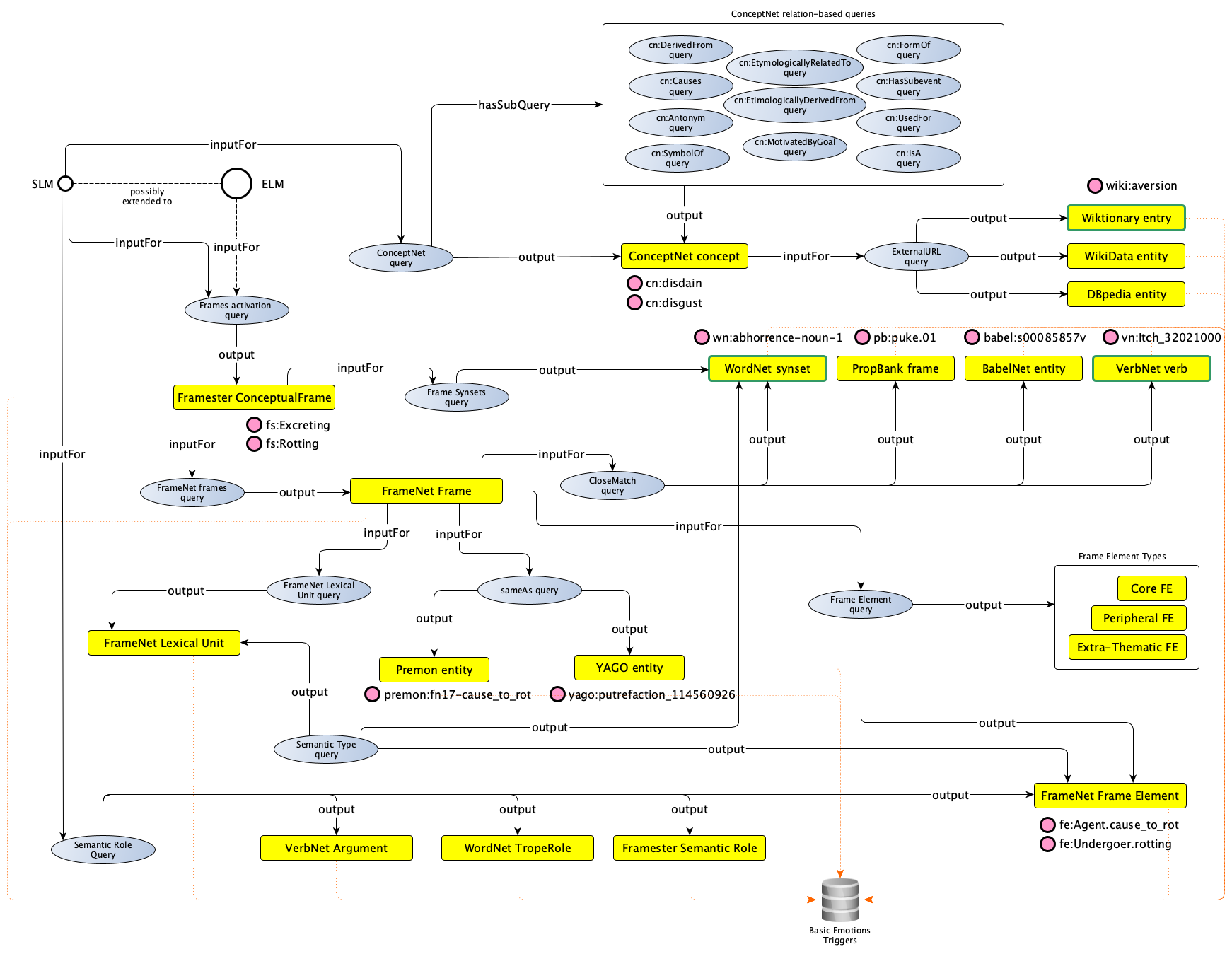}
  \caption{A knowledge graph population workflow with semantic triggers for \texttt{be:Disgust}, extracted from the Framester hub: FrameNet frames and frame elements, WordNet synsets, VerbNet verbs, DBpedia entities, Wikidata lexical entries, ConceptNet concepts, BabelNet entities and PropBank frames. Yello rectangles are classes of entities, while blue ovals are SPARQL queries, provided as additional material in: \url{https://anonymous.4open.science/r/EFO-C124}}
  \label{fig:quokka_emotions}
\end{sidewaysfigure}

The BE module follows the frame semantics idea that a frame can be evoked by a lexical unit. For example, ``rejoice'', ``serenity'', and ``appreciate'' are understood with respect to an ``Enjoyment'' conceptual frame.
To exemplify its usage, let's consider the \texttt{be:Disgust} frame. The same process has been performed for each \texttt{be:BE\_Emotion}.
To operationalize EFO and the BE frames, we generated a rich knowledge graph of entities triggering an emotion, reusing lexical and factual resources. To do so, we queried the Framester ontology \cite{gangemi2016framester}, described in Sect. \ref{sec:be}, to gather entities from different semantic web resources.
Fig. \ref{fig:quokka_emotions} shows the querying workflow, described in the following.
Finally, Table BE\_stats shows details about entities retrieved as triggers from each resource, for each emotion\footnote{Table BE\_stats is available here: \url{https://anonymous.4open.science/r/EFO-C124/BE\_Stats.png}}.

The \texttt{be:Disgust} class has 7 subclasses, here ranked in increasing intensity: Dislike $>$ Aversion $>$ Distaste $>$ Repugnance $>$ Revulsion $>$ Abhorrence $>$ Loathing\footnote{The ``$>$'' symbol is used here instead of repeating the \texttt{be:moreIntenseThan} object property}. These lexical units constitute the Starting Lexical Material (SLM) for the \texttt{be:Disgust} frame. 
Each unit of the \texttt{be:Disgust} SLM set is used as input to perform query expansions, and retrieve entities from the semantic web resources in the Framester hub. As shown in Fig. \ref{fig:quokka_emotions}:
\begin{itemize}
    \item Frame triggering gathers: \texttt{fscore:Excreting}, \texttt{fscore:BeingRotted}, \newline \texttt{fscore:CauseToRot} and \texttt{fscore:Rotting} FrameNet frames for \texttt{be:Disgust};
    \item Frame-element-driven triggering: the \texttt{be:Disgust} frame inherits the frame elements from FrameNet frames both as triggers and as its own frame elements: \newline
    \texttt{fe:Manner.CauseToRot}, \texttt{fe:Undergoer:CauseToRot} and \texttt{fe:Place.CauseToRot}; 
    \item Lexical triggering: the \texttt{be:Disgust} frame is declared \texttt{efo:triggeredBy} by all the entities subsumed by the extracted frames, e.g., WordNet synsets and VerbNet verb senses such as \texttt{wn:synset-putrefactive-adjectivesatellite-1}, \newline \texttt{wn:synset-putrefy-verb-1}, \texttt{vn:Putrefy\_45040000};
    \item skos:closeMatch triggering: the extracted frames provide a bridge to other entities having  a close match to them, e.g.: \texttt{yago:rancidity\_114561839}, \texttt{premon:fn17-excreting}, \texttt{pb:puke.01} and \texttt{babel:s00028852n} (``muck'') respectively from YAGO, Premon, PropBank and BabelNet resources;
    \item ConceptNet triggering: the SPARQL query expansion on the ConceptNet side, as shown in Fig. \ref{fig:quokka_emotions}, performed for each degree of intensity of \texttt{be:Disgust}, gather triggers such as \texttt{cn:dislike} and \texttt{cn:disdain}.
\end{itemize}

\subsection{EFO Evaluation}
\label{sec:emotion_detection}

A graph-based emotion detector for EFO has been implemented on top of the FRED knowledge extractor \cite{gangemi2017semantic}, \cite{DBLP:journals/semweb/GangemiPRNDM17}. FRED can be considered as a ``situation analyzer'': it performs hybrid knowledge extraction from natural language, based on both statistical and rule-based components, which generate RDF/OWL knowledge graphs, including entity linking, word-sense disambiguation, and frame/semantic role detection. Being directly linked to the Framester ontology, its graphs include: (i) word sense disambiguation to the WordNet resource, (ii) VerbNet verb sense disambiguation, including their semantic roles;
(iii) frame detection from FrameNet; (iv) PropBank frame recognition; (v) DBpedia entity linking.
FRED is available as an online service and as an API\footnote{\url{http://wit.istc.cnr.it/stlab-tools/fred/demo/}}.

The graph-based detection process is composed of the following three main steps: 
\begin{enumerate}
    \item A sentence in natural language is given as input to FRED tool. FRED parses the sentence and builds a knowledge graph of semantic dependencies, performing frame extraction, WordNet disambiguation, entity recognition, etc.
    \item 
    For each Framester node in the graph, we perform a SPARQL query to the BET graph, to extract emotion triggers
    \item For each emotion trigger, a triple is added to the original graph declaring which emotion is triggered.
\end{enumerate}

We tested the graph-based detector on the WASSA 2017 dataset \cite{mohammad2017wassa}, a corpus generated in the context of the 8th Workshop on Computational Approaches to Subjectivity, Sentiment and Social Media Analysis. 
This dataset was chosen because, unlike many others, is annotated on each single sentence, and not e.g. for an entire dialogue. 

The WASSA2017 corpus provides 4 datasets, annotated with 4 of Ekman’s basic emotions: fear, anger, enjoyment and sadness. Each dataset consists of approximately 1000 tweets, all labeled with one of the emotions.

\begin{figure}[hp]
    \centering
    \includegraphics[width=1\textwidth]{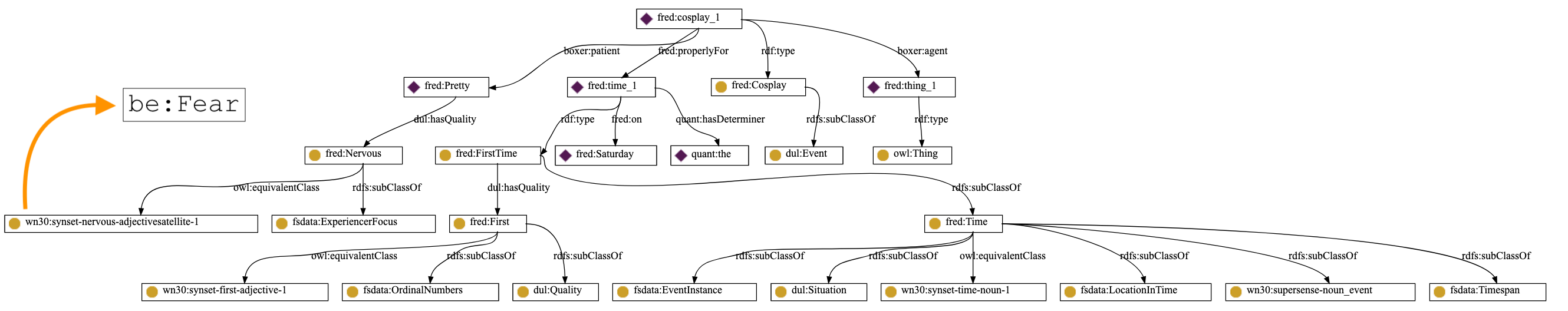}
    \caption{FRED graph generated from the sentence ``Cosplaying properly for the first time on Saturday! Pretty nervous...'' showing the localization of the activation of the \texttt{be:Fear} emotion frame on the WordNet synset \texttt{wn:nervous-adjectivesatellite-1}.}
    \label{fig:fear_fred}
\end{figure}

An example of a FRED graph enriched with emotion annotation is shown in Fig. \ref{fig:fear_fred}.

\begin{table}[]
\centering
\begin{tabular}{|l|l|l|l|l|}
\hline
          & Precision & Recall & F1 Score & Pearson Correlation \\ \hline
Anger     & 100       & 34.6   & 51.41    & 0.60                \\ \hline
Fear      & 100       & 34.76  & 51.59    &  0.59                \\ \hline
Sadness   & 100       & 26.23  & 42.06    & 0.64                \\ \hline
Enjoyment & 100       & 28.41  & 44.25    & 0.60                \\ \hline
\end{tabular}
\caption{WASSA 2017 Precision, Recall, F1 and Pearson Correlation score with frame-based emotion detector.}
\label{tab:detection}
\end{table}

We summarize here the results for each emotion frame used in WASSA2017 annotations.

\begin{itemize}
    \item \textbf{Enjoyment}: the Enjoyment WASSA2017 file contains 824 tweets. Out of these 824 tweets, only in 711 cases FRED successfully produces a graph, due to shortness of the sentence, or irregular syntax. Out of these 711, we detect \texttt{be:Enjoyment} in 204 occurrences, \texttt{be:Surprise} in 67, and  combinations of various emotions in less than 10 occurrences each;
    \item \textbf{Sadness}: the sadness WASSA2017 file is composed by 786 tweets, out of which FRED generates graphs for 752 cases. In 200 cases it is retrieved some \texttt{be:Sadness} evocation. An additional confusing factor is the fact that each sentence in the dataset is annotated with only one emotion, even when it clearly contains more than one, such as graph 501, generated from ``The immense importance of football is sometimes scary. When you don’t win you are responsible for so many unhappy people. - Arsene Wenger''. For this sentence the detector retrieves \texttt{be:Sadness} and \texttt{be:Fear};
    \item \textbf{Anger}: The anger WASSA 2017 file is composed by 857 tweets, out of which 816 graphs are generated, for a total amount of 282 \texttt{be:Anger} evocations. It seems that there is a correlation between the evocation of \texttt{be:Anger} and the FrameNet frame \texttt{fs:EmotionHeat}\footnote{The \texttt{fs:EmotionHeat} frame represents the semantics around those manifestations of emotions related to body temperature, it includes situations usually from the ``anger'' and ``love-arousal'' domains.}, in particular triggered by situations involving the WordNet \texttt{wn:blood-noun-1}, and the VerbNet \texttt{vn:Fume\_31030800} and \texttt{vn:Boil\_45030000} entities;
    \item \textbf{Fear}: The Fear file is the biggest among the WASSA 2017 files, composed by 1157 tweets. FRED generates graphs for 1076 cases. The activation of \texttt{be:Fear} is retrieved in 374 occurrences. The occurrences of other basic emotions and basic emotions combinations are shown in Fig. \ref{fig:fear}. There seems to be a certain correlation between \texttt{be:Sadness} and \texttt{be:Fear}, in particular around the notion of ``despair''. 
\end{itemize}

Many sentences result as not containing any emotion content, and this is probably due to their brevity, irregular syntax, or use of ironic language.
We have then measured whether our knowledge graph-based detection correlates with WASSA2017 annotations. Since WASSA2017 is a `bronze' standard (labels have been derived automatically from Twitter hashtags, and manually evaluated on a small selection), the precision, recall and F1 measures reported in Table \ref{tab:detection} are only indicative \emph{as if} WASSA2017 were a gold standard. The Interrater agreement correlation measure is more appropriate, because it compares two different approaches (WASSA2017 bronze standard and EFO knowledge graph-based detection). Interrater agreement using Pearson $r$  show moderate agreement, which may suggest a minor bias in the detection methods: hashtags can be motivated by content, opinion, or even irony; frame-based detection could be affected by contrasting frames. It could also suggest an inadequacy in the choice of the categorical model, or its labeling, since (as reported in \cite{coppini2022frameco}), real life affective situations can be much richer than what a simple categorisation could express.

\noindent All the graphs produced during the frame-based emotion detection are available on the EFO GitHub repository\footnote{https://anonymous.4open.science/r/EFO-C124/WASSA2017.zip}.

\section{Multimodal EFO Application}
\label{sec:multimodal_efo}

Finally, EFO is designed to handle multimodal datasets, and to show how this can be done we transposed into RDF format and integrated into the ontology network two resources available online: (i) the CREMA-D dataset\footnote{CREMA-D dataset is available here: \url{https://github.com/CheyneyComputerScience/CREMA-D}} \cite{cao2014crema} (Crowd-sourced Emotional Multimodal Actors Dataset), and (ii) the FER+ dataset\footnote{The FER+ dataset is available here: \url{https://github.com/Microsoft/FERPlus}} \cite{barsoum2016training} (Facial Emotion Recognition). We used these datasets for three main reasons: (i) they are fully available online; (ii) they explicitly adopt (versions of) Ekman's theory; (iii) they both adopt a multi-labeling approach, meaning that each item (audio clip or image) is not restricted to a single emotion label, and both datasets keep a score of the ``Emotion profile'' composition, transposed in RDF format as follows.


\paragraph{\textbf{CREMA-D}}
This datasets consists of 7442 audio recorded by 91 actors, speaking the same sentence with various intonations, according to 6 emotional states: Anger, Disgust, Fear, Happy, Neutral, and Sad.
The experiment then collected 219,687 annotation from 2,443 participants annotating the above mentioned audio clips acording to the emotion perceived.
We named CREMA-kg the RDF transposition of CREMA-D.
Each audio clip is introduced in the CREMA-kg as an \texttt{efo:EmotionSituation}, represented as a \texttt{owl:NamedIndividual}, named with the ID of the file. Each file is represented as a situation in which occur one or more emotions.
For each emotion annotated whose value is $>$ 0, two kinds of triples are generated: (i) the first one declaring the emotion value: \texttt{cr:1001\_DFA\_ANG\_XX} \texttt{fer:hasAngerValue} \texttt{"5.0"}\^{}\^{}\texttt{xsd:float}; (ii) the second one linking the file to the \texttt{be:Emotion}: \texttt{cr:1001\_DFA\_ANG\_XX} \texttt{be:includesSignalOf} \texttt{be:Anger}, \texttt{be:Disgust}.
This allows to retrieve audio clips with a specific numerical value for more than one emotion, according to their emotion profile.
The CREMA-kg knowledge graph is available on the EFO GitHub repository\footnote{The CREMA-kg is available here: \url{https://anonymous.4open.science/r/EFO-C124/CREMA-kg.ttl}}.

\paragraph{\textbf{FER+}}
This dataset is built as an improvement of the original FER dataset \cite{zahara2020facial}, it consists of 35.887 images of people's faces, each image is tagged by 10 annotators labeling the facial expression as a manifestation of one emotional state among the following: neutral, happiness, surprise, sadness, anger, disgust, fear, contempt, and unknown. We called FER-kg its RDF transposition.

\begin{figure}
    \centering
    \includegraphics{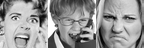}
    \caption{A sample result obtain via SPARQL querying  the FER-kg module. Here some examples of \texttt{:EmotionSituation}, including some facial expressor of emotions with values major than "3". From left to right: \texttt{fer:0035509} including \texttt{be:Anger} and \texttt{be:Fear}; \texttt{fer:0014735} including \texttt{be:Enjoyment} and \texttt{be:Anger}; and \texttt{fer:0030527} including \texttt{be:Sadness} and \texttt{be:Disgust}.} 
    \label{fig:fer_img}
\end{figure}

We excluded the 'unknown' label because it is just a confirmation of the limits of committing to a unique theory (in this case Ekman's basic emotions) for emotion detection tasks, which is one of the main reasons for the existence of EFO. For the same reason we excluded 'contempt', since it is a hybridization of Basic Emotions theory with the Contempt-Anger-Disgust models \cite{rozin1999cad}, not yet introduced in EFO.
Each image is considered as an \texttt{efo:EmotionSituation}, represented as a \texttt{owl:NamedIndividual}, named with the ID of the image. Each image is represented as a situation in which occur one or more emotions.
For each emotion annotated whose value is $>$ 0, two kind of triples are generated: (i) the first one declaring the emotion value: \texttt{fer:0014735} \texttt{fer:hasAngerValue} \texttt{"3.0"}\^{}\^{}\texttt{xsd:float}; (ii) the second linking it to the \texttt{be:Emotion}: \texttt{fer:0014735} \texttt{be:includesSignalOf} \texttt{be:Fear}.
This allows to query images with a specific value for more than one emotion, combining them as shown in Fig. \ref{fig:fer_img}.
The FER-kg graph is available on the EFO GitHub repository\footnote{https://anonymous.4open.science/r/EFO-C124/FER-kg.ttl}.

\noindent The full ontology will be available online and queryable from a permanent SPARQL endpoint at camera-ready time.

\section{Conclusions}

We have presented EFO, the Emotion Frame Ontology, which provides a formal ontology to integrate existing emotion theories and annotated multimodal datasets. We have considered the different perspectives on emotion situations (physiological states, behavioral and expression patterns, social practices, linguistic categorisation, appraisal, etc.) assumed in emotion theories, and we have designed a frame-based ontology network to formalise those theories, and make them interoperate in the context of broad-range emotion situations.
We have exemplified the design, automated reasoning and application of EFO with respect to  Ekman's Basic Emotions (BE) Theory. An evaluation has been presented by lexicalizing 
BE emotion frames from within the Framester knowledge graph, and implementing a graph-based emotion detector from text. We have also described the extension of knowledge graph-based emotion detection on emotional speech and emotional face expressions datasets.

%
%
\bibliographystyle{splncs04}
%
\bibliography{bib}
\end{document}